\documentclass[letterpaper, 10 pt, conference]{ieeeconf}  

\IEEEoverridecommandlockouts                              
\usepackage{cite}

\usepackage{amsmath,amssymb,amsfonts}
\usepackage{algorithmic}
\usepackage{graphicx}
\usepackage{textcomp}
\usepackage{xcolor}
\usepackage{comment}
\usepackage{etoolbox}
\usepackage{siunitx}
\usepackage{booktabs}
\usepackage[svgnames]{xcolor} 

\makeatletter

\patchcmd{\thebibliography}
  {\settowidth\labelwidth{\@biblabel{#1}}}  
  {\settowidth\labelwidth{\@biblabel{99}}}  
  {}{}  
\makeatother
\def\BibTeX{{\rm B\kern-.05em{\sc i\kern-.025em b}\kern-.08em
    T\kern-.1667em\lower.7ex\hbox{E}\kern-.125emX}}
\begin{document}

\title{
A Novel Bio-Inspired Fish Robot with Tunable Stiffness via Particle Jamming
 }

\author{Jack Stonecipher$^{1}$, Allen Gao$^{1}$, and Wei Wang$^{1,*}$
\thanks{$^{1}$Marine Robotics Lab, Department of Mechanical Engineering, University of Wisconsin--Madison, Madison, WI 53706, USA.}%
\thanks{$^{*}$Corresponding author: Wei Wang (e-mail: wwang745@wisc.edu).}
}

\maketitle 

\begin{abstract}
Fish achieve efficient swimming across varied speeds through active modulation of their body flexibility. To explore the effects of tunable stiffness on swimming performance, we present a bio-inspired freely-swimming fish robot with a rapidly tunable particle jamming body. This design enables rapid stiffness adjustments with negligible changes in shape or volume, achieving a 54\% variation in flexural rigidity across vacuum pressures of 0 to –40 kPa. We visualize the midline of the oscillating body under both low and high stiffness conditions, and the comparison confirms that the body curvature varies with stiffness. We further experimentally evaluate the tunable stiffness body's effects on swimming performance using velocity and cost of transport (CoT) measurements obtained via a motion tracking system. Results show that active stiffness tuning is essential for sustaining efficient and high-speed swimming across beating frequencies of 1–3 Hz. At low frequencies (1-1.5 Hz), a softer body (0 kPa) maximizes velocity and minimizes CoT, whereas at high frequencies (2.5-3 Hz), a stiffer body (–40 kPa) delivers superior velocity and reduced transport cost. These findings highlight stiffness modulation as a key strategy for adaptive and efficient propulsion in bio-inspired robotic swimmers.
\end{abstract}


\section{Introduction}
Over millions of years, fish have evolved and refined locomotor strategies that allow rapid and efficient movement through aquatic environments \cite{videler1993}. Among these adaptations, body flexibility is described as a hidden axis of biomechanical diversity in fish \cite{IJimenez_2023}. Flexibility directly impacts thrust generation and is thus one of the fundamental determinants of swimming performance.



Previous studies have demonstrated that fish actively adjust their body stiffness to regulate swimming speed and efficiency. 
For example, to increase swimming speed, fish increase body stiffness through muscle activation \cite{schwalbe_red_2018}.
More specifically, Schwalbe et al. \cite{schwalbe_red_2018} observed that bluegill sunfish (\textit{Lepomis macrochirus}) achieve this stiffening via co-activation, simultaneous left–right lateral red muscle activation at the same anteroposterior location. Co-activation persisted for longer portions of the swimming cycle during acceleration compared to steady, low-speed swimming \cite{schwalbe_red_2018}. 
However, maintaining high stiffness requires sustained muscle activity, which is energetically costly \cite{Sch1_Muller_2006, Sch2_Webb_1992, Sch3_Bale_2014}. Consequently, fish instinctively adjust stiffness and flexibility to maximize propulsive efficiency under varying conditions \cite{SCARADOZZI2017437, IJimenez_2023}.

\begin{figure}[!htb]
\centerline{\includegraphics{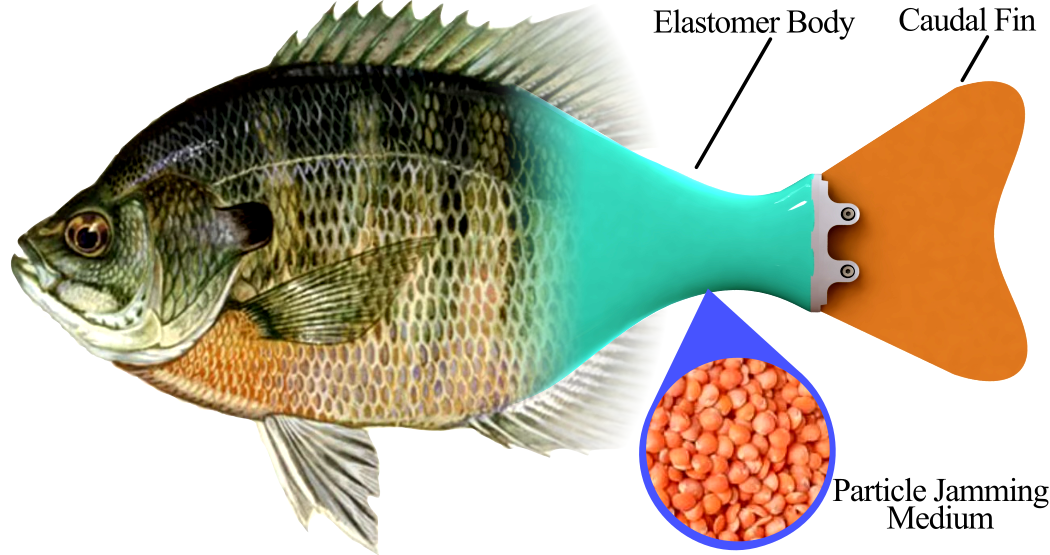}}
\caption{Bluegill-inspired fish robot with particle jamming body. 
}
\label{splashImage}
\end{figure}



As a novel class of underwater vehicles, robotic fish hold considerable potential for diverse applications \cite{zhao2025bio, zhai2025interpretable}. Since MIT's development of the first robotic fish, RoboTuna, in 1994 \cite{roboTuna}, numerous other designs have been proposed and developed \cite{currentStatusofBiomemeticRobotFish} \cite{Prakash_2024}. Yet robotic fish swimming performance still lags significantly behind that of real fish. Bridging this gap has become a central challenge in bio-inspired robotics. A prevailing strategy is to transition robotic fish structures from rigid to flexible designs, a direction supported by extensive experimental validation \cite{softMarineRobotReview2025, softMarineRobotReview2024,lu2023}. For example, a soft robotic fish, SoFi, was developed to observe and swim alongside marine life \cite{sofi}. 

More recently, researchers have begun exploring variable-stiffness mechanisms and robots as a means to further enhance the swimming performance of flexible robotic fish  \cite{Quinn_2022}.
Jusufi et al. \cite{jusufi2017undulatory} introduced a bilateral pneumatic actuator for robotic fish, enabling stiffness modulation by varying the internal cavity pressure. Similarly, Chen et al. \cite{chen2021body} developed a robotic fish capable of locally adjusting body stiffness via a variable-stiffness tensegrity joint and investigated how different stiffness distributions influence swimming performance. Zhong et al. \cite{zhong_2021} conducted a comprehensive study using a tuna-shaped robotic body to examine how fish adjust stiffness to enhance swimming speed and efficiency. In their design, a spring connected to the caudal fin was used to modulate joint stiffness. These studies not only provide novel insights for advancing robotic fish design but also deepen our understanding of how variable stiffness can enhance swimming efficiency. However, many studies on variable-stiffness methods for robotic fish have primarily focused on an individual joint. When stiffness adjustment is applied to the entire fish body, it will cause substantial changes to the external geometry, which can significantly affect the swimming performance. Therefore, it is essential to investigate approaches for modulating stiffness over a larger area while preserving the external form of the robotic fish.

Soft robotic fish designs have explored achieving full-body tunable stiffness. For example, Ju and Yun \cite{JU_2023} developed a hydraulically actuated fish that offers extreme flexibility. However, the stiffness-tuning mechanism would cause morphological deformation and slow response times. Such limitations hinder the robot’s ability to rapidly adjust stiffness for optimal swimming performance.

An alternative method for achieving tunable stiffness is jamming-based technology. By applying a vacuum to a membrane filled with particles, fibers, or sheets (Fig.~\ref{jammingMethod} illustrates the particle-jamming process), a structure can rapidly transition from flexible to rigid with minimal shape change \cite{MANTI2016}.

Previous studies have incorporated jamming mechanisms into fish fins but not into full-body swimming robots.
For example, Hong et al. \cite{layerJammingFish2024} employed layer jamming to increase the stiffness of a bio-inspired fish tail by an order of magnitude. However, this approach is primarily effective only for laterally compressed morphologies. Moreover, the layer-jamming fishtail was tested only in a fixed experimental setup, lacks integration into a full-body swimming robot, and has not yet been validated on freely swimming platforms.
 Cheng et al. \cite{CHENG2021} fabricated a caudal fin that combined pneumatically actuated rays with particle-jamming elements adhered to a thin membrane. Thrust tests conducted in a fixed experimental setup demonstrated that stiffness adjustment can enhance swimming performance at different frequencies. Nevertheless, real fish generate propulsion through coordinated body and caudal fin movements while swimming freely.
A free-swimming robot with full-body, jamming-based tunable stiffness is therefore needed to further investigate how stiffness modulation influences swimming performance under realistic, unconstrained conditions.

In this work, we present a bio-inspired robotic fish with a flexible particle-jamming body (Fig.~\ref{splashImage}) that enables rapid stiffness modulation with minimal changes in shape or volume. This design overcomes the limitations of previous methods by providing large-area stiffness variation while maintaining consistent external morphology.
We experimentally characterize the robot’s stiffness modulation capabilities and analyze their effects on static undulation. Performance evaluations across a range of stiffness–frequency combinations further demonstrate how tunable stiffness can be leveraged to optimize swimming speed and efficiency under diverse operating conditions.
The key contributions of this work are:
\begin{itemize}
\item The design and fabrication of a tethered, free-swimming robotic fish that uses particle jamming to achieve a tunable-stiffness body.
\item Experimental validation showing that tunable stiffness influences both speed and efficiency across a wide range of tail-beat frequencies.
\item Demonstration of precise stiffness control capable of rapid response with minimal morphological change.
\end{itemize}

\section{Method}
In this section, we describe the design of the bio-inspired bluegill fish robot, explain the working principle of the particle-jamming system, and detail the fabrication of the tunable-stiffness body. 

\subsection{Robot Design}
\begin{figure}[t]
\centerline{\includegraphics[width=1\linewidth]{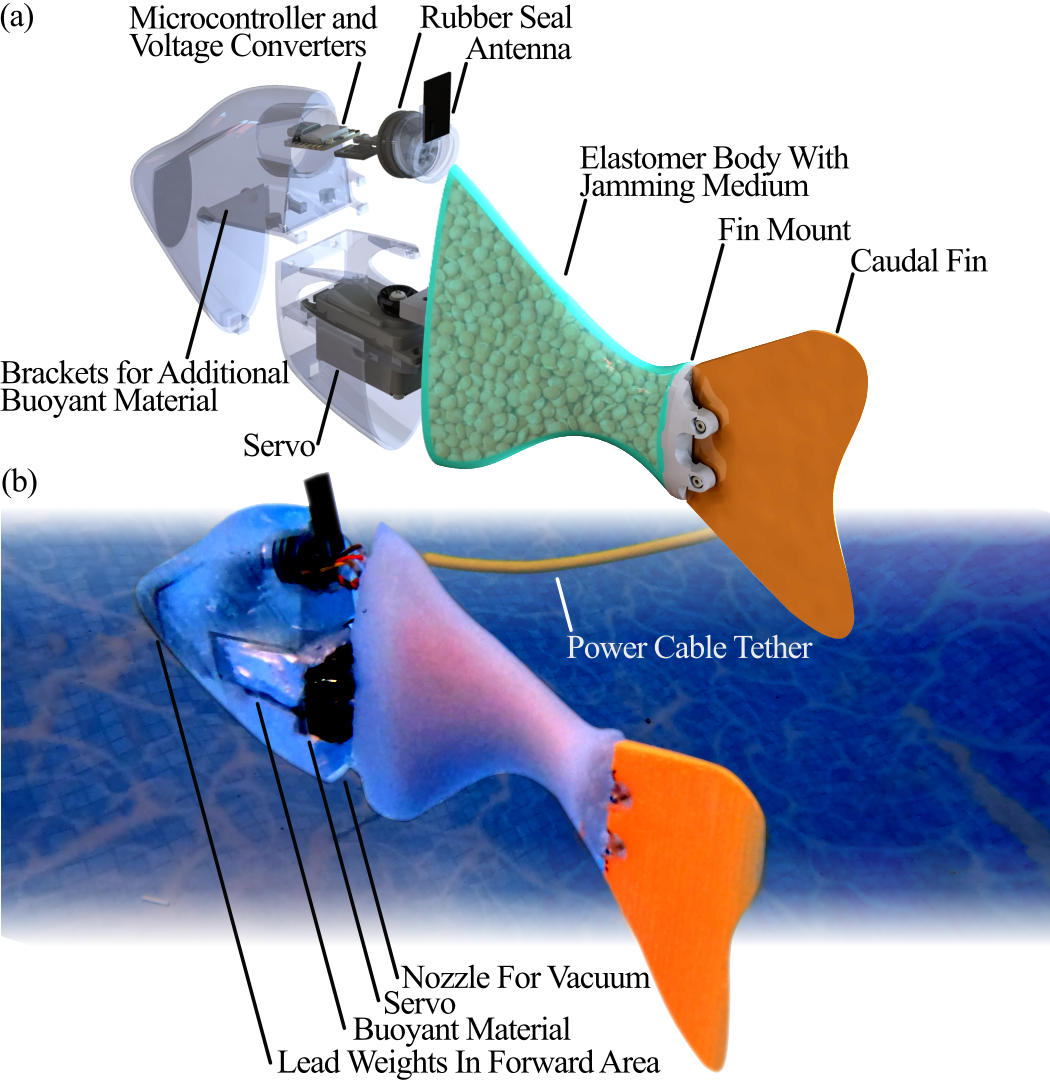}}
\caption{Design of robot fish with jamming body. (a) The robot fish's internal components. Wires routed through the plug are sealed with epoxy. The rubber seal prevents water ingression during swimming. Dimensions of the robot are  292$\times$40 $\times$123 mm (L$\times$W$\times$H) (b) Photo of fish swimming in water with a buoyancy neutral power tether cable}
\label{design}
\end{figure}

The robotic fish’s locomotion is driven by an IP67-rated waterproof servo (DB778WP, Hitec RCD USA) that connects the rigid head to the soft elastomer body (Fig.~\ref{design}). The servo follows a sinusoidal waveform to generate undulatory motion, propelling the robot through the water. The head is firmly attached to the servo support structure using four M2 screws secured to threaded inserts (Hapric) embedded within recesses on the head. The head is hollow in order to host several necessary components.

Internal electronics housed in the robotic fish’s head provide both autonomous swimming functionality and remote-control capability. The system features a XIAO ESP32C6 microcontroller (Seeed Studio) for generating PWM signals to drive the servo, supported by two voltage converters: a PW-D Control Buck Converter (WolfWhoop) supplying 5 V to the microcontroller and an MH-MINI360 converter (Teyleten Robot) delivering 7.4 V to the servo. Swimming parameters including amplitude, frequency, and steering offset are wirelessly adjustable via the ESP-NOW protocol, which enables communication between the onboard ESP32 and a computer-connected ESP32. All electronics remain sealed within the head using a rubber plug mechanism. 

A resin-printed plug provides a waterproof interface allowing wired connections between internal and external components, including the servo and a power tether connected to a 8 volt power supply. The head is sealed using a hydraulic syringe rubber stopper (30 mL disposable syringe, Omawrt) to prevent water ingress, while an external antenna mounted on the plug ensures stable wireless communication. Epoxy potting (B0BKR2HDM3, Promise Epoxy) secures the wire channels to further guard against leakage. The waterproof seal both protects electronics and traps air for buoyancy.

Buoyancy and weight distribution were optimized for stable swimming dynamics. The sealed head functions as a buoyant air chamber, with additional buoyancy available via styrofoam inserts around the servo or on external brackets. To generate a sufficient moment into the body for effective propulsion, 37 g of lead weight was positioned in a forward compartment of the head. Swimming is performed by oscillating the head and the tunable stiffness jamming body with a servo joint.

\subsection{Particle Jamming Principle}

\begin{figure}[tbp]
\centerline{\includegraphics{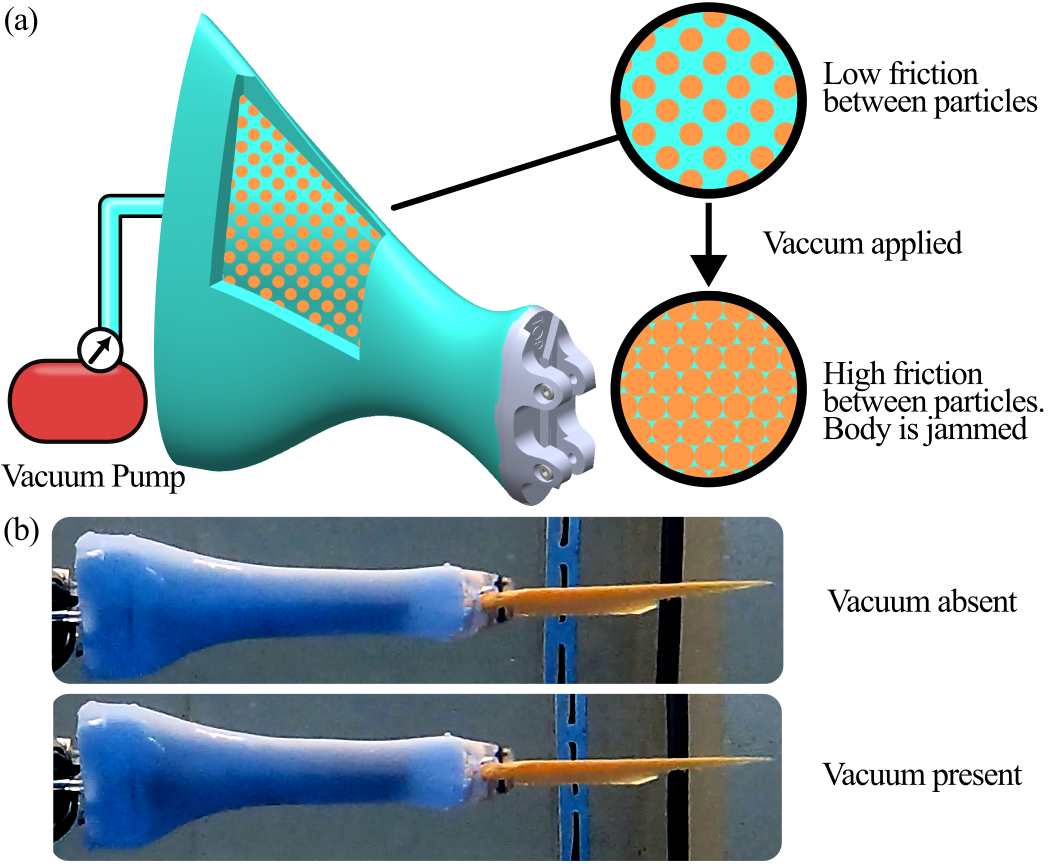}}
\caption{Particle Jamming Principle. (a) Particles (orange) inside the body cavity can move freely with no vacuum present. The presence of a vacuum causes the membrane to constrict and stiffen the body. (b) Body shape is not significantly altered with and without the presence of a vacuum}
\label{jammingMethod}
\end{figure}

Particle jamming is a variable-stiffness mechanism that uses a membrane filled with granular material, which contracts around the particles when a vacuum is applied (Fig.~\ref{jammingMethod}). The increased friction between particles dramatically stiffens the structure. 

As a stiffening mechanism, particle jamming presents many advantages. A key benefit of this method is its ability to achieve significant stiffening with minimal volume change \cite{MANTI2016}. A well-known example is the universal particle-jamming gripper developed by Brown et al. \cite{BROWN2010}, which can conform to and grasp objects while exhibiting only about $0.5\%$ volume change. Similarly, our particle-jamming body shows minimal visible difference between jammed and unjammed states (Fig.~\ref{jammingMethod}b). Another benefit of particle jamming is the ability to transition rapidly between soft and rigid states: stiffening under vacuum can be quickly reversed by applying positive pressure \cite{BROWN2010,MANTI2016}. For these reasons particle jamming is an ideal method for a tunable stiffness body, but other considerations were made while implementing it for a freely swimming robot fish.

\begin{figure}[htbp]
\centerline{\includegraphics{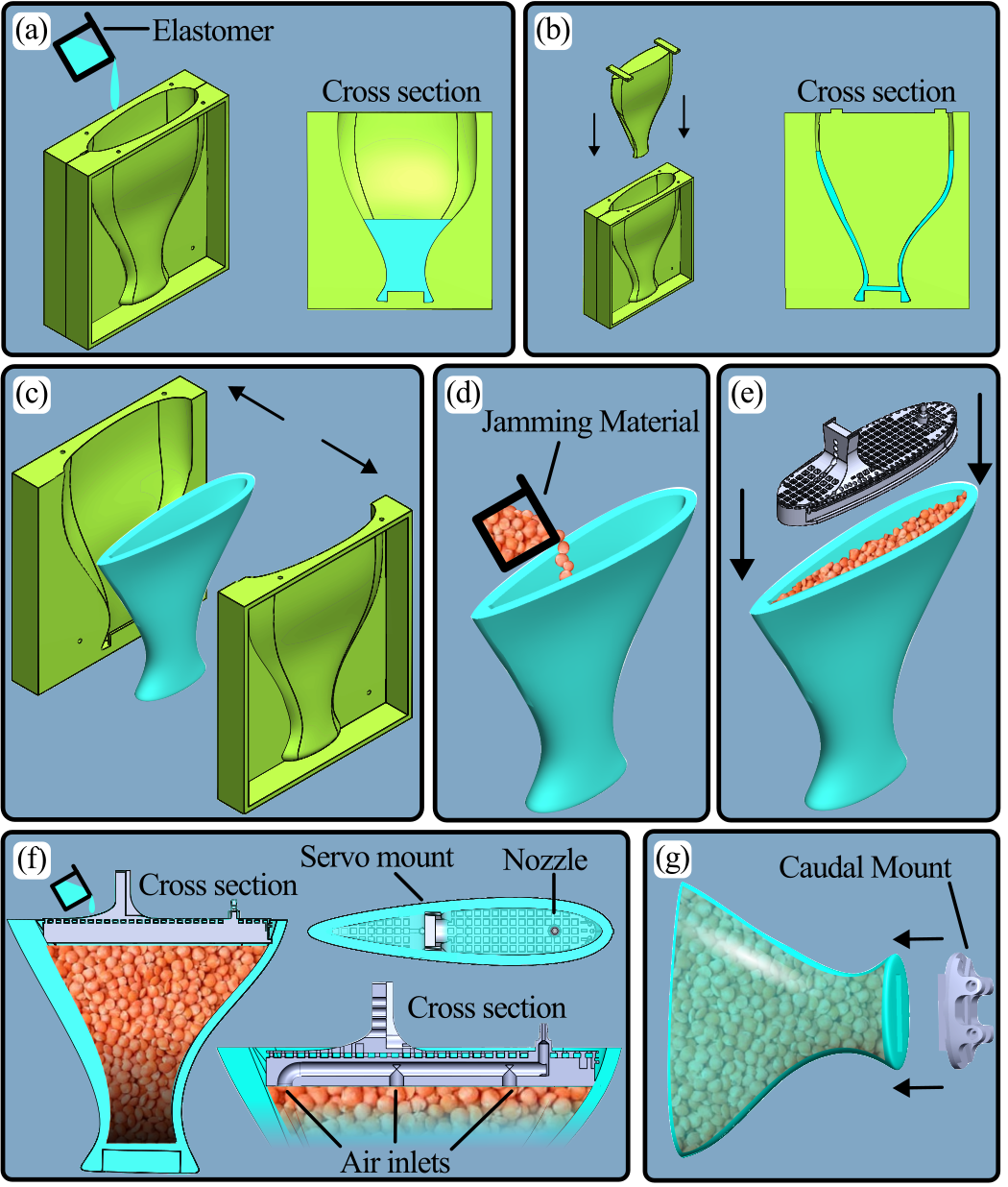}}
\caption{Fabrication process for the tunable stiffness particle jamming body. (a) Uncured elastomer is poured into two clamped molds before degassing. (b) Center mold is inserted before curing. (c) The cured elastomer is carefully removed from the three mold pieces. (d) Granular jamming medium is added to the body cavity. (e) Solid mount is inserted into the opening of the cavity  (f) Elastomer is poured over the solid mount and seeps into a lattice structure while degassing. After curing, the solid mount is physically bonded with the elastomer, creating an airtight seal around the cavity. (g) Finally, a solid caudal fin mount is inserted into a notch in the elastomer body and bonded with a clear sealant.}
\label{fabrication}
\end{figure}

For our tunable-stiffness body, selection of the jamming medium was guided by three factors—particle weight, size, and buoyancy. Dried red lentils were chosen as they meet these criteria: their low weight allows the head to exert sufficient moment on the body for effective swimming, their slightly negative buoyancy prevents undesirable pitch, and their relatively large size enables a large body volume while keeping overall weight low. 
The elastomer body is bonded to a solid mount that both secures it to the servo and incorporates internal air inlets, allowing a vacuum to be applied to the membrane (Fig.~\ref{fabrication}f).

\subsection{Fabrication of Particle Jamming Body}

The fabrication of the jamming body utilizes standard soft robotics elastomer casting techniques. A three-piece mold was printed using Clear V4 Resin (Form 4, Formlabs Inc.), washed in isopropyl alcohol for 10 minutes, and soaked in water for 30 seconds. After air drying, the print was cured in a Formlabs UV curing station for 30 minutes at 60 °C. The two mold sections were then fastened with M2 nuts and bolts, and an additional sealing layer was applied along the seams to prevent leakage. For extra protection against elastomer leakage during the casting process, a layer of sealing tape was applied over the seams.

We used a flexible and soft elastomer with a shore hardness of 00-30 (Ecoflex 00-30, Smooth-On inc.) to expand the range of stiffness achievable by the particle jamming body. We poured 80 g of mixed elastomer (Ecoflex 0030; 40 g Part A and 40 g Part B) into the mold (Fig.~\ref{fabrication}a). The elastomer mixture was degassed under a vacuum of -14 psi for 10 minutes to remove trapped air bubbles, preventing voids, porosity, or surface defects. After degassing, an additional 40 g of uncured elastomer was mixed and added to compensate for volume loss during degassing. The center section of the mold was then inserted and screwed into place to create the cavity for the jamming material (Fig.~\ref{fabrication}b). The filled mold was degassed again at -14 psi for 5 minutes before being baked at 60 °C for 35 minutes. After curing and being removed from the mold (Fig.~\ref{fabrication}c), excess elastomer was trimmed away and 35 g of dried red lentils (the jamming medium) were inserted into the internal cavity (Fig.~\ref{fabrication}d). 

An SLA resin–printed servo mount and nozzle assembly, designed with internal airways for vacuum application, was inserted into the cavity and clamped into place temporarily (Fig.~\ref{fabrication}e). A layer of elastomer was poured over the solid mount (Fig.~\ref{fabrication}f), which featured a lattice structure allowing the elastomer to flow into and bond with the part \cite{howtoMoldSilicone}. The assembly was degassed for 15 minutes to remove air pockets from the lattice, then baked for 35 minutes at 60 °C, positioned upright to ensure a flat bond line. Once cured, the elastomer bonded not only to the previously cast layer but also integrated firmly with the SLA-printed component. This seal must be airtight against both water pressure and internal applied vacuum, since water ingression into the body would compromise the tuning mechanism and the buoyancy of the robot fish. The connection of the tail to the servo and head must also be strong. The lattice bond technique achieves a  connection that would be difficult to replicate using adhesives or mechanical fasteners \cite{howtoMoldSilicone}. The elastomer body is entirely soft, save for this lattice piece and the caudal fin mount. 

The caudal fin mount, designed to be lightweight and nearly neutrally buoyant, connects the fin to the elastomer body. This resin-printed “peduncle” component (Fig.~\ref{fabrication}g) is attached using a clear sealant (732 Clear Multipurpose Sealant, Dow Chemical Company). The forked caudal fin, inspired by Bluegill sunfish, is fabricated from thermoplastic urethane (TPU 85A, Bambu Labs) with a tapered thickness profile, providing greater flexibility at the tip than at the base. This flexibility allows the swimming motion of the elastomer body to continue into the caudal fin.

\section{Results}
In this section we describe the testing methods and results characterizing the stiffening capabilities of the fish. We then analyze the effect stiffening has on the traveling wave formed by a body being oscillated by a mounted servo. Finally we explore the impressive effect of stiffness on swimming performance metrics, including velocity and efficiency.

\subsection{Stiffness Characterization}
\begin{figure}[htbp]
\centerline{\includegraphics{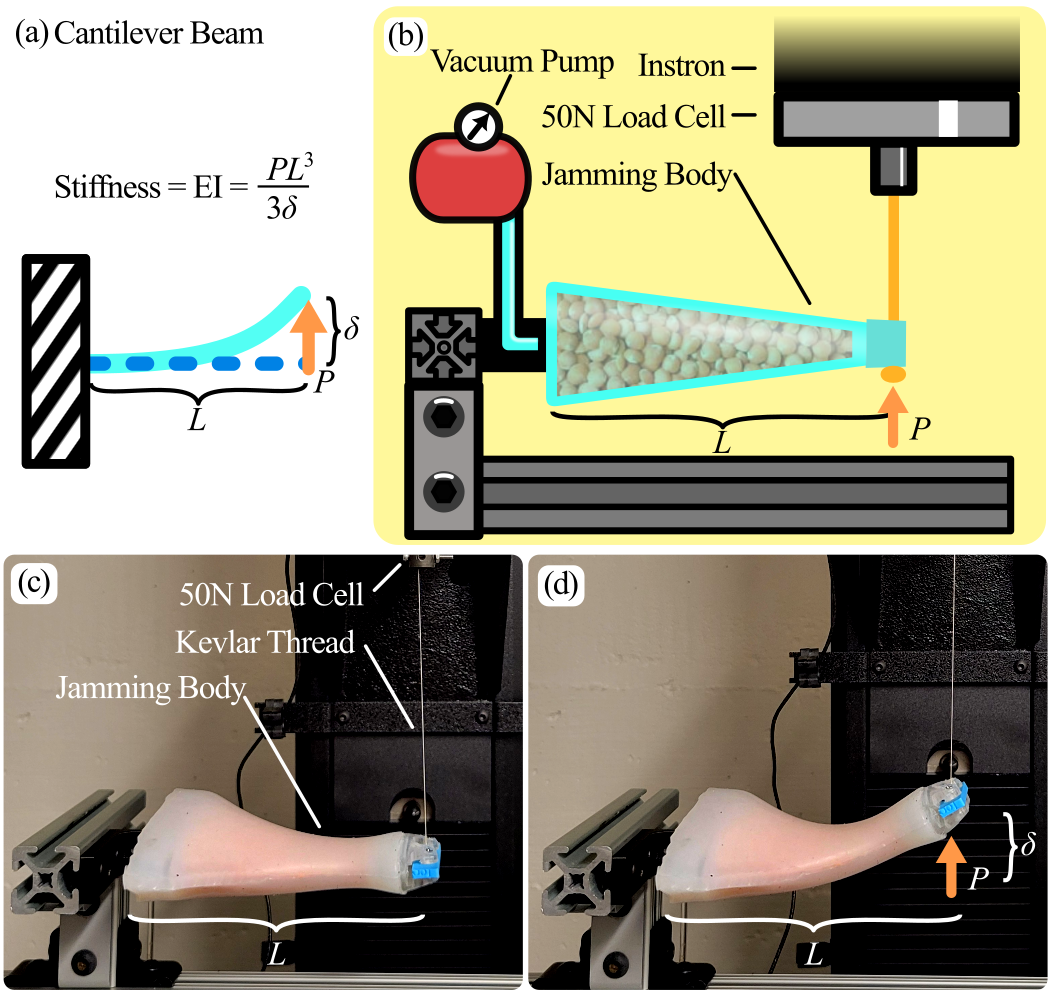}}
\caption{Testing apparatus characterizing the correlation between vacuum level and particle jamming body stiffness. (a) Theoretical model of a cantilever beam described by equation \ref{beamEq} (b) Illustration of an apparatus to measure the correlation between effective stiffness and vacuum pressure. The Instron records the resistance to deformation force from a tip displacement of 25 mm (c) Photo of the jamming body before displacement, tip connected to the load cell by Kevlar thread. The load cell set to 0 N at this position. (d) The Instron moves, as programmed, to a displacement of 25 mm and records the resulting point force \textit{P}.}
\label{stiffnessSetup}
\end{figure}

To quantify the relationship between vacuum level and body stiffness, we modeled the robotic fish body as a cantilever beam subjected to a point load (Fig.~\ref{stiffnessSetup}a). By displacing the tip of the jamming body by a known amount and measuring the resulting reaction force, we calculated the body’s equivalent flexural rigidity, \(EI\)~(N$\cdot$m$^{2}$), which characterizes its stiffness. This value was obtained using the beam deflection equation:  
\begin{equation}
EI = \frac{P L^{3}}{3 \delta}
\label{beamEq}
\end{equation}
where \(EI\) is the flexural rigidity (N$\cdot$m$^{2}$), \(P\) is the applied point load (N), \(L\) is the beam length (m), and \(\delta\) is the tip displacement (m). To ensure accurate stiffness measurements, we designed an apparatus capable of precisely displacing the body tip while simultaneously recording the corresponding reaction force.

We attached an Instron tensile tester (68SC-1, Instron Tool Works Inc.) equipped with a 50~N static load cell (2530-50N, Instron Tool Works Inc.) to the free end of the elastomer jamming body, while securing the servo-connection end to the base of the Instron (Fig.~\ref{stiffnessSetup}b).  
The Instron was programmed to move a specified distance and record the resistance force at a peak displacement of 25~mm (displacement error of $\pm0.01$~mm).  
For each stiffness level, the average force (error of $\pm0.1$~N) was calculated from five repeated trials.  
The body was stiffened using a diaphragm vacuum pump (Huanyu) at vacuum levels between $-20$~kPa and $-40$~kPa.  
Lower vacuum levels were generated using a hydraulic syringe coupled with a pressure gauge. Even small amounts of vacuum were enough to alter the stiffness of the particle jamming body.

\begin{figure}[htbp]
\centerline{\includegraphics{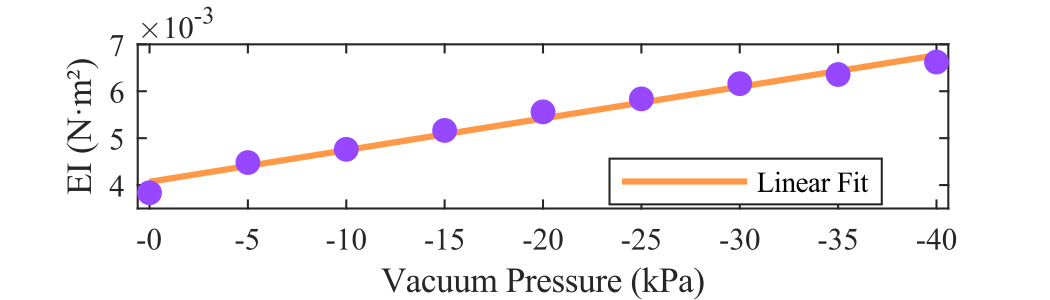}}
\caption{Body stiffness under vacuum pressures ranging from 0 to -40 kPa, averaged over 5 trials for each stiffness level.  The line of best fit is EI = 0.000067 * Vacuum + 0.004070, with an R$^{2}$ value of 0.982. Standard deviation of collected data was between 0.00002 and 0.00007.}
\label{stiffnessResults}
\end{figure}

Our results (Fig.~\ref{stiffnessResults}) show that the stiffness of the particle-jamming body can be tuned approximately linearly by adjusting the vacuum pressure. The stiffness increases $54\%$ from a baseline of $0.0038~\mathrm{N \cdot m^{2}}$ with no vacuum to $0.00662~\mathrm{N \cdot m^{2}}$ at $-40~\mathrm{kPa}$. This broad, predictable stiffness range is critical for an effective tunable-stiffness fish robot, as it allows the device to optimize its swimming performance across a wide spectrum of beating frequencies. Altering the stiffness can have pronounced effects on the interaction between the body and the water surrounding it.

\subsection{Wave Propagation Under Varied Stiffness}

\begin{figure}[htbp]
\centerline{\includegraphics[width=0.9\linewidth]{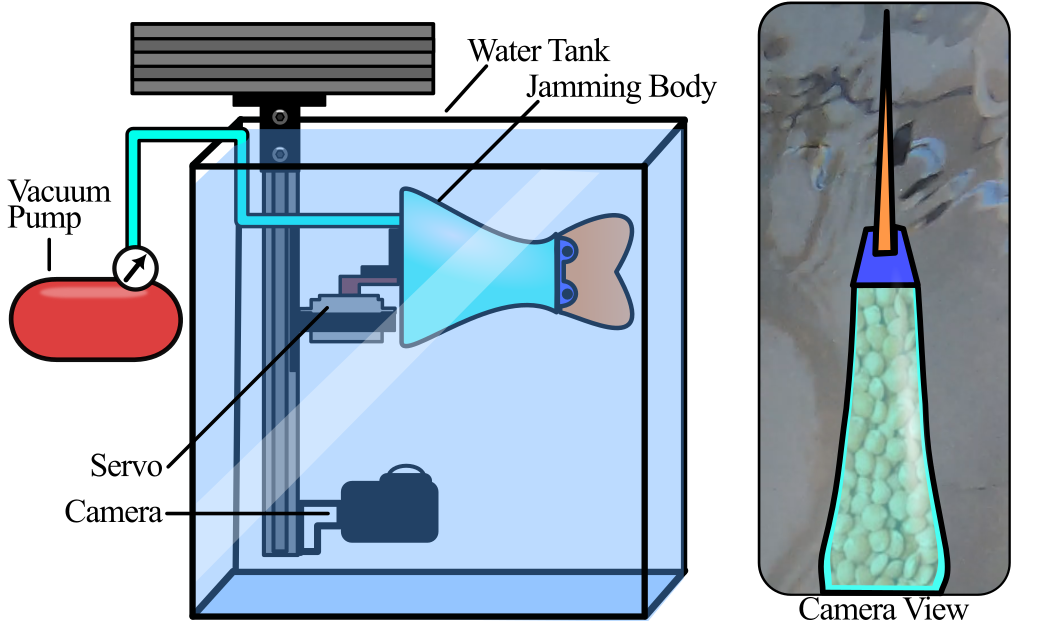}}
\caption{Photographic setup to record body undulation changes under different stiffness levels. Servo and jamming body mounted above a camera. The camera footage is processed to overlap body wave propagation from the same beating frequency at different levels of stiffness.}
\label{oscillationSetup}
\end{figure}

To examine how body stiffness influences the traveling wave generated by servo-driven undulatory motion, we mounted the robotic body and servo on an aluminium structure (1010 series, 80/20) and positioned a stationary camera (GoPro Hero 9, GoPro Inc.) below to capture the whole movement of the body during oscillation (Fig.~\ref{oscillationSetup}). We captured 4K video at 30~fps of body oscillations at a 15\textdegree{} amplitude across frequencies ranging from 0.5~Hz to 3~Hz. The procedure was repeated under three vacuum conditions: 0~kPa, --20~kPa, and --40~kPa. The recordings were then aligned and overlaid using video editing software (Vegas Pro 15.0, MAGIX Software) to visualize the effect of stiffness on the resulting body undulations.

\begin{figure}[htbp]
\centerline{\includegraphics{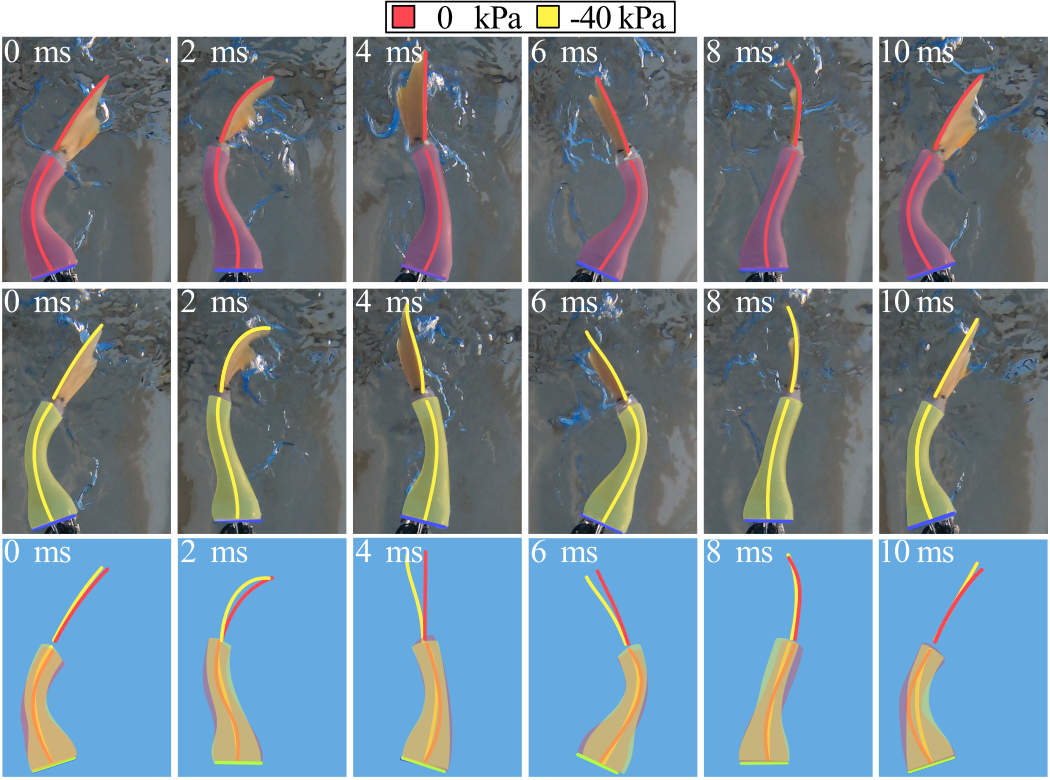}}
\caption{Comparison of undulation under 3 Hz beating frequencies with and without a vacuum present (yellow and red, respectively). Each frame is 2 ms apart. The differences in both body and fin curvature is apparent when overlaid with each other.}
\label{oscillationResults}
\end{figure}

Tunable stiffness enables the robot to regulate its hydrodynamic interactions during swimming, which directly affects locomotion performance. As shown in Fig.~\ref{oscillationResults}, comparing two bodies oscillating at 3~Hz, the stiffer body resists hydrodynamic forces more effectively and maintains a straighter midline. In contrast, the softer body exhibits pronounced bending. Such control over wave propagation is critical, as swimming performance strongly depends on how undulatory motion travels along the body \cite{roboTuna}.

\subsection{Velocity and Efficiency}
\begin{figure}[htbp]
\centerline{\includegraphics[width=0.9\linewidth]{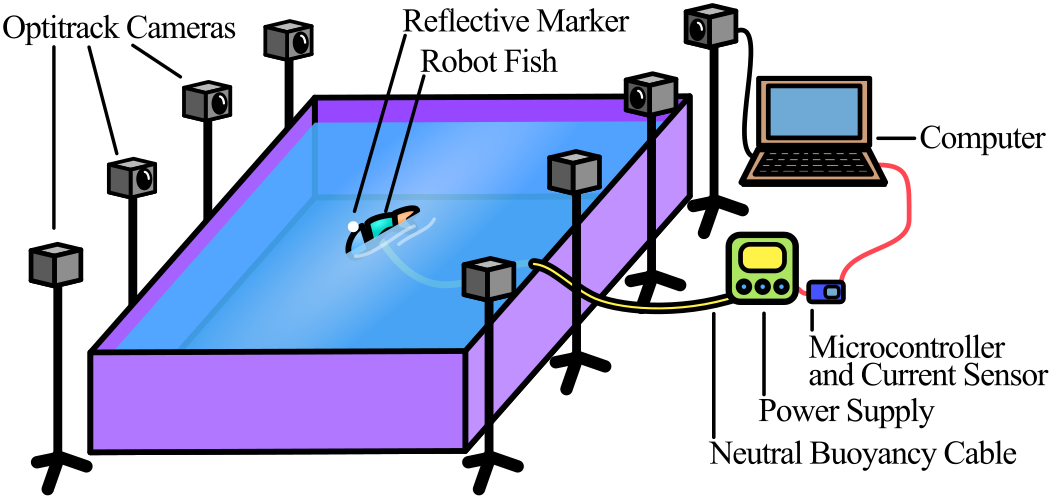}}
\caption{Motion tracking system. Eight OptiTrack cameras surround a pool to track a marker mounted to the robot fish. Positional data is synchronized with power consumption to measure velocity and cost of transport.}
\label{optitrackSetup}
\end{figure}

To evaluate the impact of the tunable stiffness system on swimming performance, we tracked the robot’s speed using an optical motion capture system (OptiTrack Prime X 13, NaturalPoint Inc.) (Fig.~\ref{optitrackSetup}). 
Operating at a sampling rate of 125~Hz with an accuracy of $\pm0.2$~mm, the system recorded the robot’s position during swimming. A reflective marker was mounted on top of the fish to maintain consistent visibility by at least three of the eight OptiTrack cameras. For each stiffness level, vacuum was applied to the jamming body through an air-channel nozzle and sealed with a tube cap. This allowed us to test with just one tether that supplied a constant 8-volt source delivered via a neutrally buoyant tether (Fathom ROV Tether, Blue Robotics).

An off-board ESP32 connected to an INA219 breakout board (Adafruit Industries) measured current draw and power consumption. This same unit remotely commanded the onboard ESP32 to adjust the robot’s beating frequency. A Python script recorded position data streamed from OptiTrack’s Motive software. Swimming speed was computed as the position change between consecutive time steps. Motion tracking over water presents challenges, as light reflections on the water’s surface can be misidentified as the robot’s marker. Large, abrupt spikes in calculated speed were flagged as misidentifications and removed using a filtering process. The filtered motion-tracking data was synchronized with the power consumption measurements. The ESP32 controlling the fish and measuring power consumption sampled at 20~Hz. Averaging results over intervals of consistent, high-quality tracking data allowed us to gain a clear picture of not just the velocity of the fish but its efficiency as well. 

Swimming efficiency was quantified using the cost of transport (CoT), defined as  
\[
\text{CoT} = \frac{P}{v},
\]  
where \(P\) is the power consumption (W) and \(v\) is the swimming velocity (m/s). CoT represents the energy required to move the robot a unit distance. Locomotion efficiency is affected by both the beating frequency of the robot and its body stiffness.

Using our power consumption and motion tracking system we measured the tunable stiffness fish robot's free swimming performance across beating frequencies from 1--3~Hz and vacuum pressures of 0, --20, and --40~kPa. Frequencies below 1~Hz produced insufficient forward thrust, while frequencies above 3~Hz caused the servo to struggle to reach the full 35\textdegree{} amplitude (70\textdegree{} peak-to-peak). Depending on frequency and body stiffness, swimming performance could vary greatly.

\begin{figure}[htbp]
\centerline{\includegraphics{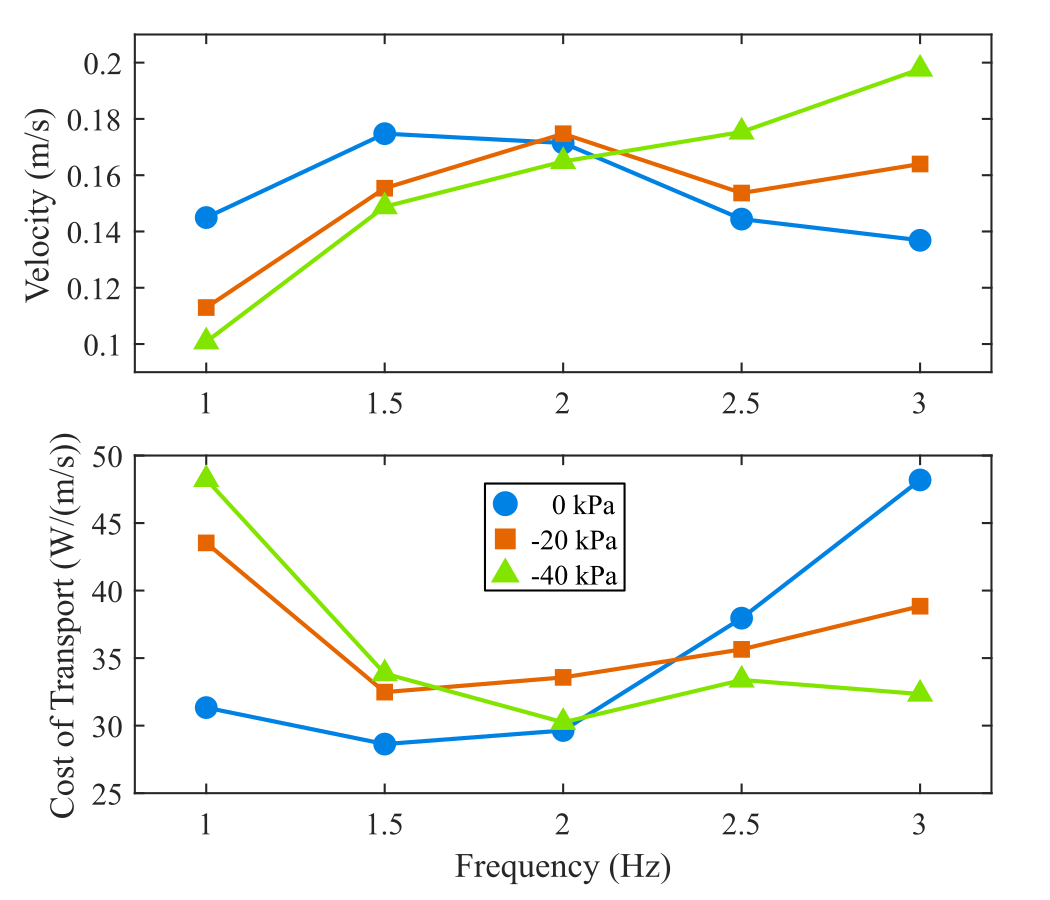}}
\caption{Velocity and Cost of transport performance. Optimal velocity and efficiency performance at lower frequencies produced by the softest body ($0~\mathrm{kPa}$, shown in blue), whereas the optimal velocity and efficiency at higher frequencies is produced with the stiffest body ($-40~\mathrm{kPa}$, shown in green).}
\label{optitrackResults}
\end{figure}

The results (Fig.~\ref{optitrackResults}, Tables~\ref{tab:velocity} and~\ref{tab:cot}) emphasize the critical role of tunable body stiffness in maintaining optimal swimming performance across a broad range of beating frequencies.
\begin{table}[!ht]
\centering
\caption{Average velocity (m/s) at different beating frequencies and vacuum pressures. Stiffness configuration with superior velocity at each frequency highlighted}
\begin{tabular}{l *{3}{S[table-format=1.5]}}
\toprule
{Frequency (\si{Hz})} & {\SI{0}{kPa}} & {\SI{-20}{kPa}} & {\SI{-40}{kPa}} \\
\midrule
1.0 & \colorbox{lightgray}{0.145} & \colorbox{white}{0.113} & \colorbox{white}{0.101} \\
1.5 & \colorbox{lightgray}{0.175} & \colorbox{white}{0.155} & \colorbox{white}{0.149} \\
2.0 & \colorbox{white}{0.171} & \colorbox{lightgray}{0.175} & \colorbox{white}{0.165} \\
2.5 & \colorbox{white}{0.144} & \colorbox{white}{0.154} & \colorbox{lightgray}{0.175} \\
3.0 & \colorbox{white}{0.137} & \colorbox{white}{0.164} & \colorbox{lightgray}{0.198} \\
\bottomrule
\end{tabular}
\label{tab:velocity}
\end{table}
\begin{table}[!ht]
\centering
\caption{Average Cost of Transport (\si{W/(m/s)}) at different beating frequencies and vacuum pressures. Stiffness configuration with superior efficiency at each frequency highlighted }
\begin{tabular}{l *{3}{S[table-format=2.3]}}
\toprule
{Frequency (\si{Hz})} & {\SI{0}{kPa}} & {\SI{-20}{kPa}} & {\SI{-40}{kPa}} \\
\midrule
1.0 & \colorbox{lightgray}{31.3} & \colorbox{white}{43.5} & \colorbox{white}{48.2} \\
1.5 & \colorbox{lightgray}{28.6} & \colorbox{white}{32.5} & \colorbox{white}{33.8} \\
2.0 & \colorbox{lightgray}{29.6} & \colorbox{white}{33.6} & \colorbox{white}{30.2} \\
2.5 & \colorbox{white}{38.0} & \colorbox{white}{35.6} & \colorbox{lightgray}{33.4} \\
3.0 & \colorbox{white}{48.2} & \colorbox{white}{38.8} & \colorbox{lightgray}{32.3} \\
\bottomrule
\end{tabular}
\label{tab:cot}
\end{table}
At lower frequencies (1 and 1.5~Hz), the soft body configuration (equivalent beam stiffness of 0.0038~N$\cdot$m$^{2}$) achieved the highest swimming speeds and the lowest cost of transport (CoT). A performance crossover occurs around 2~Hz. At higher frequencies (2.5 and 3~Hz), the stiffest configuration (equivalent beam stiffness of 0.00662~N$\cdot$m$^{2}$, achieved at a vacuum of --40~kPa) outperformed the softer bodies. For example, at 3~Hz, the stiff body reached a speed of 0.19~m/s with a CoT of 32.2~W/(m/s), exceeding the performance of the softer configurations at the same frequency. As shown in Fig.~\ref{oscillationResults}, stiffer bodies resist bending during oscillation, transmitting greater force at high frequencies, while softer bodies exhibit greater flexibility and efficiency at low frequencies.

This trend of softer bodies performing best at low frequencies and stiffer bodies excelling at high frequencies, aligns with previous tunable-stiffness robotic fish studies \cite{zhong_2021, JU_2023}, reinforcing the importance of stiffness modulation for sustaining high performance under varying operating conditions. For example, these robotic fish can adjust stiffness allows efficient swimming at low frequencies for energy-saving patrols and high stiffness for quick maneuvers in strong currents or turbulent waters.

\section{Conclusion}
In this paper we have presented a tunable stiffness fish using a particle jamming technique. This approach allows for rapid stiffening, increasing the body's flexural rigidity by 54\% to alter the swimmers interactions with the water. The swimming performance advantages of tunable stiffness are clearly demonstrated in the optimization of velocity and efficiency over a wide range of operating conditions (Fig.~\ref{optitrackResults}).

A critical next step is the inclusion of an on onboard power and pump system to achieve full tether-less operation. Autonomy could be further enhanced by integrating a camera and other sensors, enabling the robotic fish to make its own navigational decisions. 
Moreover, a reinforcement learning algorithm could be deployed to autonomously discover optimal body stiffness parameters. This process could efficiently determine the optimal stiffness for different beating frequencies, amplitudes, turning maneuvers, and even complex escape responses such as c-turns \cite{cTurnsAndStiffness}.



\bibliographystyle{IEEEtran}
\bibliography{references}

\end{document}